\documentclass{ceurart}

\usepackage{listings}

\begin{document}

\copyrightyear{2026}
\copyrightclause{Copyright for this paper by its authors.
	Use permitted under Creative Commons License Attribution 4.0
	International (CC BY 4.0).}

\conference{SuRE'26: Workshop on Sustainability and Resource-Efficiency of Artificial Intelligence, August 17, 2026, Bremen, Germany}

\title{
CompressKV: Semantic-Retrieval-Guided KV-Cache Compression for Resource-Efficient Long-Context LLM Inference
}

\author[1]{Xiaolin Lin}[%
  email=xiaolin.lin@tu-darmstadt.de,
]
\cormark[1]
\author[1]{Jingcun Wang}[%
  email=jingcun.wang@tu-darmstadt.de,
]

\author[1]{Olga Kondrateva}[%
  email=olga.kondrateva@tu-darmstadt.de,
]

\author[2]{Yiyu Shi}[%
  email=yshi4@nd.edu,
]

\author[3]{Bing Li}[%
  email=bing.li@tu-ilmenau.de,
]

\author[1]{Grace Li Zhang}[%
  email=grace.zhang@tu-darmstadt.de,
]

\address[1]{Technical University of Darmstadt, Darmstadt, Germany}
\address[2]{University of Notre Dame, Notre Dame, IN, USA}
\address[3]{Technical University of Ilmenau, Ilmenau, Germany}

\cortext[1]{Corresponding author.}

\begin{abstract}
Long-context large language model (LLM) inference is increasingly constrained by the memory footprint and decoding cost of key-value (KV) caches, limiting sustainable deployment on resource-constrained hardware. Existing KV cache eviction methods typically apply heuristic token scoring over all heads in GQA-based LLMs. These methods ignore the different functionalities of attention heads,  leading to the eviction of critical tokens and thus degrading the performance of LLMs.
To address this issue, we propose CompressKV, a resource-efficient KV-cache compression framework for GQA-based LLMs. Instead of aggregating attention scores from all heads, CompressKV identifies Semantic Retrieval Heads (SRHs) that capture both the initial and final tokens of a prompt and semantically important mid-context evidence, and uses them to select tokens whose KV pairs should be retained.
Furthermore, CompressKV allocates cache budgets across layers according to offline estimates of layer-wise eviction error. Experiments on LongBench and Needle-in-a-Haystack show that CompressKV consistently outperforms existing KV-cache eviction methods across memory budgets. Notably, it preserves over 97\% of full-cache performance using only 3\% of the KV cache on LongBench question-answering tasks and achieves 90\% accuracy with just 0.7\% KV storage on Needle-in-a-Haystack. These results demonstrate an improved resource--performance trade-off for long-context LLM inference. Our code is publicly available at:
\url{https://github.com/TUDa-HWAI/CompressKV}
\end{abstract}


\maketitle

\section{Introduction}
Recent advances in large language models (LLMs)~\cite{openai2024gpt4technicalreport,anthropic_claude3_2024,grattafiori2024llama3herdmodels,qwen2025qwen25technicalreport,jingcun2025} have boosted their long-context processing capabilities. 
However, with the increasing length of texts,  the resulting key-value (KV) cache size grows linearly. The large KV cache 
leads to slow inference due to the attention calculation across past KV cache. In addition, the large KV cache 
requires
substantial memory storage, which creates a major bottleneck in the deployment of long-context LLMs. Therefore, effective compression of KV cache is essential for optimizing the computational efficiency and model scalability.

State-of-the-art KV cache compression focuses on  quantization, low-rank approximation, and KV cache eviction ~\cite{liu2024kivi,kang2024gearefficientkvcache,ge2024modeltellsdiscardadaptive,xiao2024efficientstreaminglanguagemodels,li2024snapkvllmknowslooking,cai2025pyramidkvdynamickvcache,yang2024pyramidinferpyramidkvcache,qin2025cakecascadingadaptivekv}. Among them, KV-cache eviction—discarding KV pairs for unimportant tokens while retaining the rest—has attracted increasing attention.

Several criteria have been proposed to identify tokens for KV-cache eviction. For example, 
StreamingLLM~\cite{xiao2024efficientstreaminglanguagemodels} retains the first and last tokens 
and neglects potentially important tokens in the middle of the prompt. 
SnapKV~\cite{li2024snapkvllmknowslooking} 
clusters recent attention scores within an observation window at the end of the prompt, either per head or per head group, to identify and retain the important tokens receiving the highest attention values. CAKE~\cite{qin2025cakecascadingadaptivekv} extends SnapKV’s method by adding the  attention variance in an observation window to the eviction score, enabling it to capture tokens whose importance fluctuates over time.

While the above criteria work well in many KV-cache eviction scenarios, they overlook head heterogeneity: all heads are weighted equally, and eviction decisions are made from aggregated attention scores (typically the sum across heads within a group).
In fact, attention heads exhibit different functionalities. 
For example, in Grouped Query Attention (GQA)-based LLMs~\cite{ainslie2023gqatraininggeneralizedmultiquery}, some attention heads, called Streaming Heads, exclusively focus on the beginning and the end of a prompt ~\cite{xiao2024efficientstreaminglanguagemodels,xiao2024duoattentionefficientlongcontextllm}. When the attention heads within a GQA group are dominated by Streaming Heads, those heads have the largest influence on KV cache eviction, resulting in only the initial and last tokens’ KV pairs being retained. As a result, crucial mid-context tokens may be evicted, degrading LLM performance.

Besides eliminating KV pairs for those unimportant tokens, 
state-of-the-art research also 
allocates specified memory budgets to layers. For example, ~\cite{xiao2024efficientstreaminglanguagemodels,li2024snapkvllmknowslooking} allocates each layer to a fixed number of KV pairs without considering layer difference. 
~\cite{yang2024pyramidinferpyramidkvcache,cai2025pyramidkvdynamickvcache,qin2025cakecascadingadaptivekv} allocates KV cache budget across layers based on attention distributions or layer-wise statistics such as attention entropy or variance, which often require additional online computation cost. Moreover, attention distributions can vary significantly across models, which limits the generalization ability and effectiveness of these allocation strategies. Orthogonally, HeadKV~\cite{fu2024headsmatterheadlevelkv} and AdaKV~\cite{feng2025adakvoptimizingkvcache} extend to head-level budget allocation.

In this paper, we observe that certain attention heads 
are capable of retrieving important tokens within the text and attending to their surrounding semantic context. We refer to these heads as Semantic Retrieval Heads. 
Motivated by this observation, we identify such  Semantic Retrieval Heads in each layer and use them to determine the crucial tokens and share a unified set of crucial token indices across all heads within that layer. 
This approach can substantially address the dominance of Streaming Heads in KV cache evictions, so that it can enhance the performance of GQA-based models.
Furthermore, we analyze the cache eviction error of each layer individually and  
introduce a layer-adaptive KV cache allocation strategy. Our contributions are as follows: 

(1) We introduce a Semantic-Retrieval–driven mechanism to address streaming-head dominance in GQA, preventing important tokens from being evicted. The identified Semantic Retrieval Heads then guide token importance and KV-cache eviction. Our experimental results demonstrate Semantic Retrieval Heads know what tokens are unimportant before generation.

(2) We estimate each layer’s compression impact by computing the Frobenius norm of the difference between its attention‐block outputs with the compressed cache and those with the full cache, during the decoding stage. Cache budgets are then proportionally assigned across layers, prioritizing layers with higher errors. Importantly, this analysis is performed offline and does not introduce any additional overhead during online inference. 

(3) CompressKV is validated on multiple LLMs using LongBench and Needle-in-a-Haystack (NIAH). 
On LongBench, CompressKV maintains over 99\% of full‐cache performance with only 19\% of  KV budget and retains 97\% of question‐answering accuracy using just 3\% of the cache. On Needle‐in‐a‐Haystack retrieval benchmark, it achieves 90\% of the baseline accuracy with only 0.7\% of KV storage.

\section{Background and Related Work}
\begin{figure}[t]
    \centering
    \includegraphics[width=0.92\linewidth]{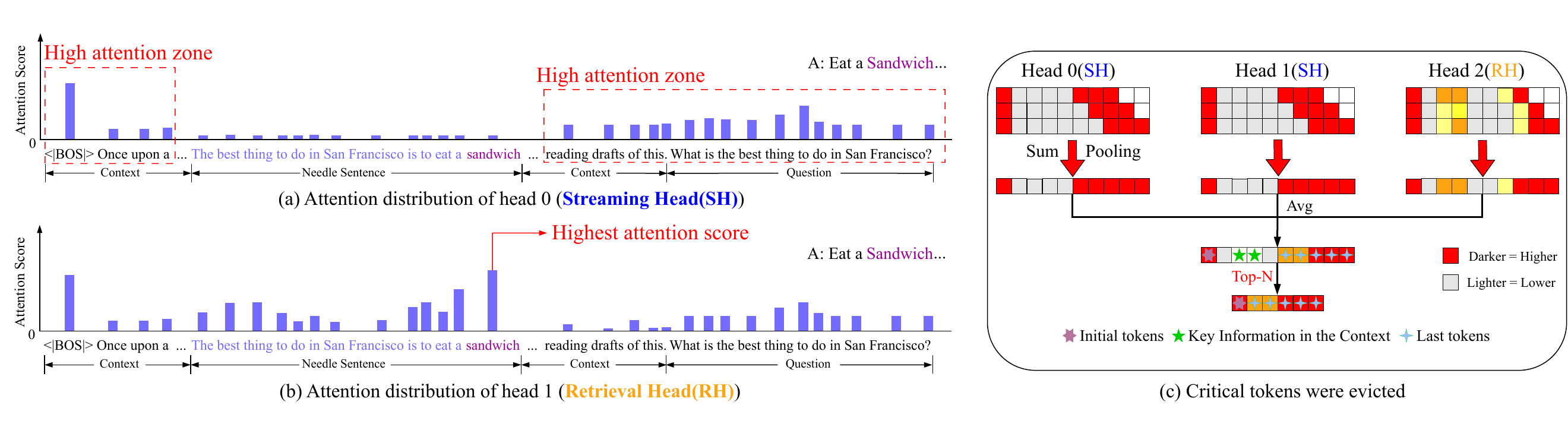}
        \caption{Motivation. (a) The attention score distribution of a streaming head (SH). (b) The attention score distribution of a retrieval head (RH). (c) Streaming attention heads in a GQA group dominate the token eviction, indicating only the initial and final tokens are retained. The critical tokens are evicted.}
    \label{fig:motivation}

\end{figure}

\subsection{KV-Cache Compression for Budgeted Long-Context Inference}
To alleviate the burden of KV cache storage, various KV cache compression methods, e.g., quantization~\cite{liu2024kivi}, low‐rank approximations~\cite{kang2024gearefficientkvcache}, and KV cache eviction strategy have been proposed. In particular,
KV cache eviction reduces cache size by removing KV cache pairs of unimportant tokens without retraining.  
There are different eviction strategies. For example, StreamingLLM~\cite{xiao2024efficientstreaminglanguagemodels} focuses solely on retaining the first and last tokens, which only addresses the Streaming Head scenario and neglects potentially important tokens in the middle of the sequence. 
To overcome this limitation, more advanced methods have been proposed~\cite{liu2023scissorhandsexploitingpersistenceimportance,zhang2023h2oheavyhitteroracleefficient,li2024snapkvllmknowslooking,han2024lminfinitezeroshotextremelength,oren2024transformersmultistaternns}. A representative example is SnapKV~\cite{li2024snapkvllmknowslooking}, which clusters recent attention scores, either per head or per head group to identify important token and retain the KV cache pairs of such tokens.
Besides, recent approaches, including PyramidKV~\cite{cai2025pyramidkvdynamickvcache},  D2O~\cite{wan2025d2odynamicdiscriminativeoperations}, and CAKE~\cite{qin2025cakecascadingadaptivekv}, dynamically allocate cache budgets based on attention statistics or modeled attention dynamics of all the layers in an LLM. 
Beyond layer-level allocation, HeadKV~\cite{fu2024headsmatterheadlevelkv} and AdaKV~\cite{feng2025adakvoptimizingkvcache} further enhance cache budget with head-level budget allocation.
Their selection strategies for important tokens are an extended version of SnapKV’s eviction strategy.

Despite their effectiveness, existing eviction pipelines have two limitations that are especially relevant to GQA-based LLMs. First, many prior KV cache eviction pipelines compute token importance via head-agnostic pooling (e.g., across heads within each GQA group) when selecting tokens for eviction, effectively treating all attention heads equally and ignoring their functional heterogeneity; Recent work~\cite{olsson2022incontextlearninginductionheads,kwon2022fastposttrainingpruningframework,zheng2024attentionheadslargelanguage,ren2024identifyingsemanticinductionheads,wu2024retrievalheadmechanisticallyexplains,todd2024functionvectorslargelanguage,yin2025attentionheadsmatterincontext,tang2024razorattentionefficientkvcache,fu2024headsmatterheadlevelkv} has shown that different attention heads have distinct roles. 
For example, some attention heads, called Streaming Heads in the state-of-the-art research, always focus on the beginning and the end of a prompt.  For example, in Figure~\ref{fig:motivation}~(a), head 0 is such a Streaming Head since the attention scores of the initial token and the last tokens are larger than the remaining tokens. 
On the contrary, some attention heads, called Retrieval heads in ~\cite{wu2024retrievalheadmechanisticallyexplains}, exhibit copy‑and‑paste behaviors for long‑context scenarios.  For example, 
in Figure~\ref{fig:motivation}~(b), head 1 is such a retrieval head since the attention scores of the correct answer ``sandwich" are larger. HeadKV~\cite{fu2024headsmatterheadlevelkv} further scores heads using retrieval and reasoning signals. In GQA-based LLMs, Streaming Heads tend to have larger effect than the other heads for KV cache eviction, which indicates only KV cache pairs corresponding to initial and last tokens are retained. This leads to the eviction of crucial tokens in the middle of a prompt and thus degrades the performance of LLMs. Figure~\ref {fig:motivation}~(c) illustrates such an example, where Streaming Heads including head0 and head1 dominate token eviction for KV cache compression. 

Second, existing layer-adaptive allocation methods~\cite{yang2024pyramidinferpyramidkvcache,cai2025pyramidkvdynamickvcache,qin2025cakecascadingadaptivekv} often rely on attention distributions or layer-wise statistics such as entropy and variance. 
These signals can introduce additional online computation and may vary across models, making the resulting allocation less robust under fixed resource budgets. In contrast, CompressKV identifies Semantic Retrieval Heads offline and uses them to guide retrieval-aware token selection, while assigning layer-wise budgets according to offline eviction-error estimates. 
This design improves the accuracy--memory trade-off without requiring online layer-importance estimation or budget search during generation.
\section{CompressKV}
CompressKV includes three key components: (1) Identification of the attention heads that are capable of retrieving important tokens within the text and attending to their surrounding semantic context. (2) Important token selection driven by such identified heads. (3) Error-aware layer-adaptive cache allocation. In the following subsections, we will first explain our observations and insights into identification of attention heads with specified functionalities. Afterwards, we will take advantage of such heads to select tokens for KV cache eviction. Furthermore, different cache budgets will be allocated to different layers.  
\begin{figure}[t]
    \centering
    \includegraphics[width=0.92\linewidth]{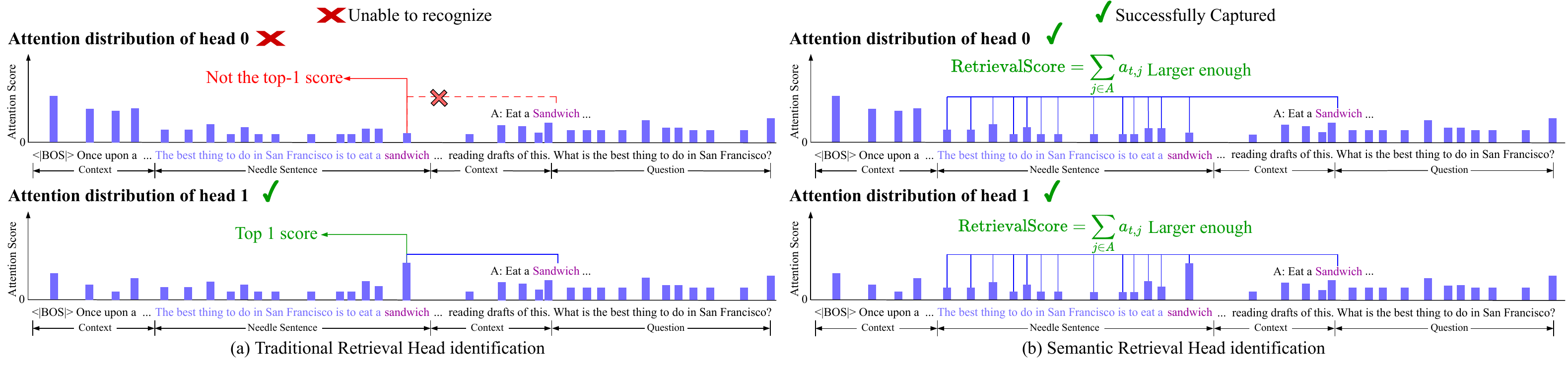}
        \caption{Illustration of Semantic Retrieval Head identification versus traditional Retrieval Head selection. Semantic Retrieval Heads capture attention over the entire answer span, addressing the limitations of traditional methods that rely solely on copy-and-paste behavior. }
    \label{fig:semantic_retrieval_head}

\end{figure}
\subsection{Observations and Insights}
To prevent Streaming Attention Heads from dominating KV-cache eviction as illustrated in Figure~\ref{fig:motivation}~(c), we use Retrieval Heads—rather than all attention heads—to identify important tokens for KV-cache eviction. Importantly, most prior methods do not leverage retrieval heads for token-level eviction decisions; for example, HeadKV mainly uses retrieval-head signals for head-level KV budget allocation instead of selecting tokens to keep or evict. To this end, we first review how prior work identifies Retrieval Heads.

Previous work identifies Retrieval Heads using a strict top-1 rule, indicating that those attention heads, the highest attention score of which aligns exactly with the correct token answer during generation, are labeled as Retrieval Heads \cite{wu2024retrievalheadmechanisticallyexplains}. This identification technique emphasizes copy-and-paste behavior.~\cite{tang2024razorattentionefficientkvcache} extends copy-and-paste identification by classifying both echo heads (copy-and-paste to the identical prior token) and induction heads (an extension that attends to the immediately preceding token) as Retrieval Heads.
HeadKV~\cite{fu2024headsmatterheadlevelkv} relaxes the strict top-1 criterion to a top-N hit: at each decoding step, a head is credited if the ground-truth answer token ranks within its top-k attention weights. 

Although HeadKV are more relaxed than strict top-1, this criteria still remains peak-driven, privileging sharp attentions on the answer token. In long contexts where attention is sparse and skewed towards boundary tokens—top-1 rules yield low hit rates and can under-credit attention heads whose attention covers the answer span and its semantic neighborhood without placing a single sharp peak on the exact answer token. In HeadKV,  if parts of the answer span do not appear within the top-k ranked positions, heads allocating substantial attention to these tokens may not be credited. For instance, in Figure~\ref{fig:semantic_retrieval_head}~(a), head 0 fails to receive credit because the relevant tokens fall outside the top-k range despite providing coverage around the correct answer. Moreover, because the top-k threshold in HeadKV is tied to the answer length, when answers are short, e.g., only one or two tokens, this method returns back to the original strict top-1 regime.

To address this limitation, we introduce Semantic Retrieval Heads (SRH), a span-aggregation standard that credits attention heads for both copy-and-paste behaviours and deeper semantic dependencies.  We then use such heads to identify important tokens for KV cache eviction, thereby preventing crucial mid-prompt evidence from being suppressed by streaming heads. For a visual comparison between Semantic Retrieval Heads and traditional Retrieval Heads, please refer to Section~\ref{sec:head_visualization}.
\subsection{Semantic Retrieval Head Identification Standards}
Instead of requiring exact top‑k hits in the traditional Retrieval Head identification, we use a calibration dataset (following \cite{wu2024retrievalheadmechanisticallyexplains}; provided in our codebase) to evaluate each head $h$ by aggregating its attention mass over the entire answer span whenever the model generates a correct answer token. Formally, we define the SRH score $S_{\text{SRH}}(h)$ as
\begin{align}
\label{formular:srh_id}
S_{\mathrm{SRH}}(h)
&= \sum_{t=1}^{N} 
\mathbf{1}_{\{y_t \in \mathcal{A}\}}
\sum_{j \in \mathcal{A}} a_{t,j}^{(h)} .
\end{align}
where $y_t$ is the generated token at step $t$, $\mathcal{A}$ is the answer span, and $a_{t,j}^h$ is head $h$’s attention weight on the $j$‑th token of $\mathcal{A}$.  The higher the score of a head is, the more capable of capturing semantic information this head is. 

Figure~\ref{fig:semantic_retrieval_head}~(b) illustrates the concept of this new identification standard. By summing over the entire span, we can capture attention heads that contribute semantically relevant context even when they never achieve top‑1 attention on a single token. Aggregation over multiple tokens enables the method to recognize heads that attend to semantic cues—such as “eat” or “a thing” around “sandwich”—rather than only pure copy‑and‑paste patterns. For example, head 0 in Figure~\ref{fig:semantic_retrieval_head} is considered as Semantic Retrieval Head in our new standard although it is not considered as Retrieval Head in the traditional identification methods. For a visual comparison between Semantic Retrieval Heads and traditional Retrieval Heads, please refer to Section~\ref{sec:head_visualization}.
\begin{figure}[ht]
    \centering
    \includegraphics[
        width=0.92\linewidth,
        height=3.5cm,
        keepaspectratio
    ]{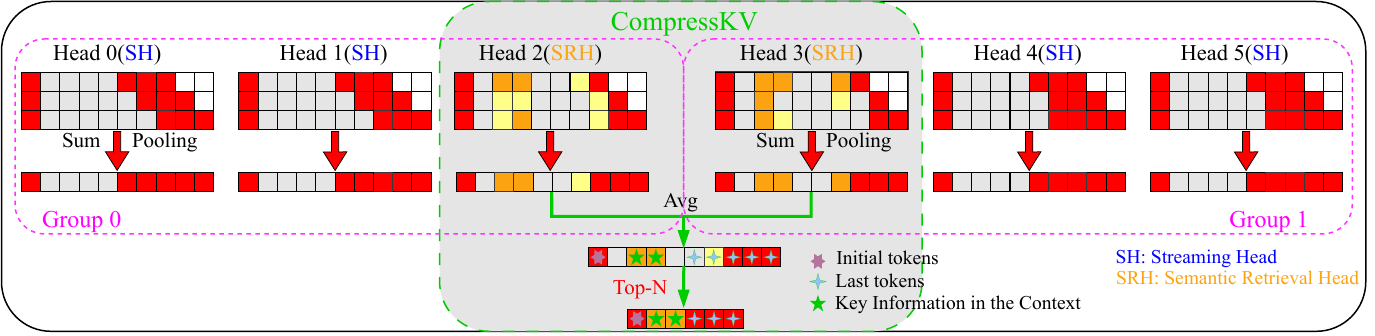}
    \caption{Illustration of the token selection driven by Semantic Retrieval Heads.}
    \label{fig:main_idea}
  
\end{figure}
\subsection{Token Selection Driven by Semantic Retrieval Heads}
In GQA-based LLMs, for each layer, we will select the top-$k$ Semantic Retrieval Heads with high scores defined with equation~\ref{formular:srh_id} as the criterion for selecting important tokens for KV cache eviction. 
All the attention heads within this layer share a common set of selected token indices determined by these top Semantic Retrieval Heads. This concept is illustrated in Figure~\ref{fig:main_idea}, where a layer has two groups. In this example, Head 2 and Head 3 are top 2 Semantic Retrieval Heads. Following SnapKV, the attention score matrices of such heads are compressed by summing over the observation window and pooling across the token dimension.
Afterwards, such compressed vectors are averaged. The tokens with the top $N$ highest attention scores will be selected and their corresponding KV cache pairs will be retained. The KV cache pairs for the remaining tokens will be evicted to compress KV cache.  
\subsection{Error-Aware Layer-Adaptive Cache Allocation}
To maximize memory efficiency under strict budget constraints, we propose an error-aware and layer-adaptive cache allocation strategy. Instead of relying on attention statistics as in the previous methods, this approach quantifies the compression error caused by KV cache compression, using full-cache outputs as the reference.
We specifically focus on the extreme compression setting, where only a small fraction of tokens are retained in each layer’s KV cache. For each layer $l$ and decoding step $t$, let $\mathbf{O}_{\text{full}, t}^l$ and $\mathbf{O}_{\text{comp}, t}^l$ denote the attention outputs using the full and compressed KV caches, respectively:
\begin{align}
\mathbf{O}^l_{\text{full},t}
&= \mathbf{W}^l_{O}\,\mathrm{Attn}\!\left(
\mathbf{Q}^l_t,\ \mathbf{K}^l_{\text{full}},\ \mathbf{V}^l_{\text{full}}
\right), \label{eq:ofull}\\
\mathbf{O}^l_{\text{comp},t}
&= \mathbf{W}^l_{O}\,\mathrm{Attn}\!\left(
\mathbf{Q}^l_t,\ \mathbf{K}^l_{\text{comp}},\ \mathbf{V}^l_{\text{comp}}
\right). \label{eq:ocomp}
\end{align}
where $\mathbf{W}_O^{(l)}$ is the output projection matrix of layer $l$, $\mathbf{Q}_t^l$ is the query, $\mathbf{K}^l$ is the key, and $\mathbf{V}^l$ is the value representation at layer $l$.
To evaluate the error incurred by compressing KV cache per layer, the error score for layer $l$ is computed and normalized as:
\begin{equation}
\label{eq:layer_error_all}
\begin{aligned}
e^{(l)}
&= \sum\nolimits_{t=1}^{T}
\frac{
\| \Delta \mathbf{O}^{l}_{t} \|_{F}
}{
\| \mathbf{O}^{l}_{\mathrm{full},t} \|_{F} + \epsilon
},
\quad
\tilde{e}^{(l)}
&= \frac{e^{(l)}}{\sum\nolimits_{k} e^{(k)}} 
\end{aligned}
\end{equation}
 where $T$ is the total number of decoding steps,$\|\cdot\|_F$ denotes the Frobenius norm and $\epsilon$ is a small positive constant (e.g., $10^{-6}$) to prevent division by zero. 

Given the normalized per-layer error scores ${\tilde{e}}$ and total cache budget $B_{total}$, 
we first assign a minimum allocation $m$ and a maximum allocation $M$ to each layer to avoid a layer either has no memory budget or a large memory budget.  
The remaining budget is distributed in proportion to the error scores. 
\section{Experiments} 
\paragraph{Baselines and Models.} 
We compare CompressKV with six KV-cache eviction baselines: 
StreamingLLM~\cite{xiao2024efficientstreaminglanguagemodels}, 
SnapKV~\cite{li2024snapkvllmknowslooking}, 
PyramidKV~\cite{cai2025pyramidkvdynamickvcache}, 
CAKE~\cite{qin2025cakecascadingadaptivekv}, 
HeadKV~\cite{fu2024headsmatterheadlevelkv}, 
and AdaKV~\cite{feng2025adakvoptimizingkvcache}. 
All methods are evaluated with greedy decoding on Llama-3.1-8B-Instruct~\cite{grattafiori2024llama3herdmodels}, Mistral-7B-Instruct-v0.3~\cite{jiang2024clipdinovisualencoders}, Qwen2.5-14B-Instruct, and Qwen2.5-32B-Instruct~\cite{qwen2025qwen25technicalreport}. 
We further examine the orthogonality of CompressKV in Section~\ref{sec:orthogonal_integration}, where we integrate it with head-level allocation, prefilling acceleration, and KV-cache quantization.
\paragraph{Evaluating Tasks.}
To evaluate CompressKV's performance under different memory budgets, we adopt two comprehensive benchmarks and one masking‑based ablation analysis:
(1) LongBench~\cite{bai2024longbenchbilingualmultitaskbenchmark}, which contains 16 long-context subtasks across single-document QA, multi-document QA, summarization, few-shot learning, synthetic tasks, and code completion. (2) Needle‑in‑a‑Haystack(NIAH)~\cite{gkamradt2024llmtest}, which measures the retrieval of a target answer hidden in extended text; 
and (3) an ablation of retrieval head types, following ~\cite{wu2024retrievalheadmechanisticallyexplains}, where we selectively disable SRH and TRH to quantify their contributions. We also compare CompressKV with TRH vs.\ SRH under equal per-layer KV budgets, e.g., 256 tokens and report results separately.
\paragraph{Implementation Details.}
We evaluate all methods under average per-layer KV-cache budgets, denoted as $B_{\mathrm{per\text{-}layer}}$, ranging from 128 to 2048 tokens. 
Given a total KV-cache budget $B_{\mathrm{total}}$ over $L$ transformer layers, 
$B_{\mathrm{per\text{-}layer}} = B_{\mathrm{total}}/L$ denotes the average budget assigned to each layer. 
StreamingLLM and SnapKV use uniform layer-wise budgets, whereas PyramidKV, CAKE, and CompressKV redistribute budgets across layers under the same total memory constraint. 
HeadKV and AdaKV are applied at the GQA-group granularity to respect grouped-query attention. 
For fairness, all methods evict tokens only during prefilling and use the same local attention setting as SnapKV~\cite{li2024snapkvllmknowslooking}: $\texttt{window\_size}=8$ and $\texttt{kernel\_size}=5$.

For CompressKV, we select the top four SRHs per layer, identified offline once per model using the calibration data from~\cite{wu2024retrievalheadmechanisticallyexplains}. 
Layer-adaptive allocation is also computed offline by measuring normalized Frobenius-norm reconstruction errors between compressed-cache and full-cache attention-block outputs under minimal-size KV compression on LongBench. 
We constrain each layer's budget to $[m,M]$, with $m=32$ and $M=3 \times B_{\mathrm{per\text{-}layer}}$, and allocate the remaining KV pairs proportionally to the normalized errors. 
During inference, the precomputed SRH sets and layer-wise budgets are fixed and reused for all samples, so CompressKV does not require online layer-importance estimation or additional dynamic profiling during generation.

\subsection{Evaluation on LongBench Benchmark} 
\begin{figure}[!ht]
    \centering
    \includegraphics[width=0.92\columnwidth]{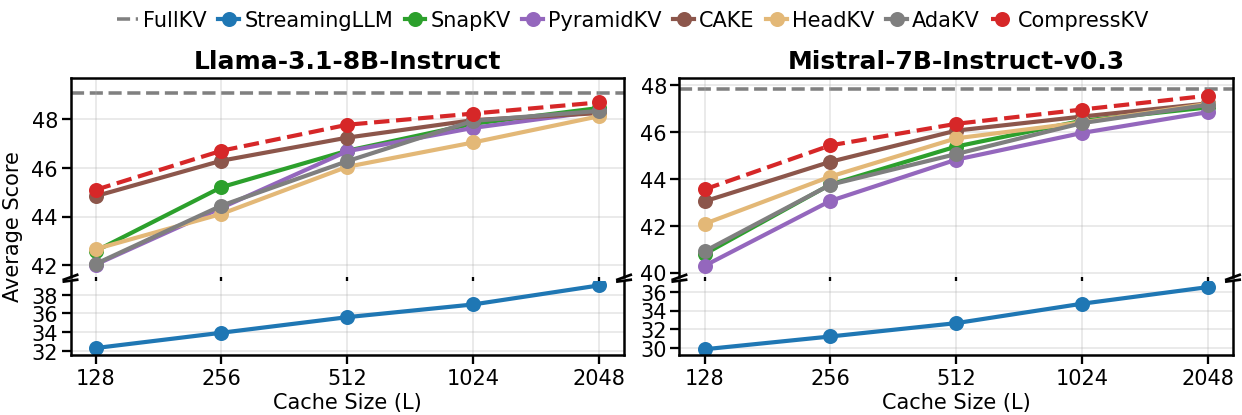}
    \caption{Average performance on 16 LongBench datasets under varying KV-cache budgets, compared with baseline methods.}
    \label{fig:longbench_cache_budgets}

\end{figure}

Table~\ref{tab:longbench_avg} reports average LongBench scores under two representative KV-cache budgets: 256 tokens for tight memory settings and 1024 tokens for moderate compression. CompressKV consistently achieves the highest average performance across model families and scales, ranging from Llama-3.1-8B-Instruct(Llama-3.1-8B) and Mistral-7B-Instruct-v0.3(Mistral-7B) to larger Qwen2.5-14B-Instruct(Qwen2.5-14B) and Qwen2.5-32B-Instruct(Qwen2.5-32B) models. The gains are most pronounced under the tighter 256-token budget, showing that CompressKV is especially effective when KV-cache memory is severely constrained. These results also indicate that the benefit of CompressKV generalizes from 7B/8B-scale models to larger 14B/32B models.
As illustrated in Figure~\ref{fig:longbench_cache_budgets}, we further benchmark CompressKV on LongBench across KV-cache sizes from 128 to 2048 using Llama-3.1-8B-Instruct and Mistral-7B-Instruct-v0.3. CompressKV consistently outperforms all baselines across the full budget range, with the largest margins at small cache sizes. These results show that SRH-guided token selection and layer-adaptive budget allocation improve the memory--performance trade-off of long-context LLM inference, especially under strict memory constraints.

\begin{table}[t]
\centering
\caption{
Average LongBench scores under fixed KV-cache budgets.
FullKV is the uncompressed reference and does not depend on the budget.
The Budget row indicates the average retained KV tokens per layer for compressed methods.
}
\label{tab:longbench_avg}
\begin{tabular}{lcccccccc}
\toprule
\multirow{4}{*}{Method}
& \multicolumn{4}{c}{Small-scale LLMs}
& \multicolumn{4}{c}{Large-scale LLMs} \\
\cmidrule(lr){2-5} \cmidrule(lr){6-9}
& \multicolumn{2}{c}{Llama-3.1-8B}
& \multicolumn{2}{c}{Mistral-7B}
& \multicolumn{2}{c}{Qwen2.5-14B}
& \multicolumn{2}{c}{Qwen2.5-32B} \\
\cmidrule(lr){2-3} \cmidrule(lr){4-5}
\cmidrule(lr){6-7} \cmidrule(lr){8-9}
& 256 & 1024
& 256 & 1024
& 256 & 1024
& 256 & 1024 \\
\midrule

\rowcolor{gray!15}
FullKV
& \multicolumn{2}{c}{49.08}
& \multicolumn{2}{c}{47.82}
& \multicolumn{2}{c}{49.80}
& \multicolumn{2}{c}{48.57} \\

\rowcolor{blue!8}
Budget
& 256 & 1024
& 256 & 1024
& 256 & 1024
& 256 & 1024 \\
\midrule

StreamingLLM
& 33.92 & 36.95
& 31.22 & 34.73
& 25.86 & 29.84
& 25.28 & 29.62 \\

SnapKV
& 45.21 & 47.82
& 43.76 & 46.48
& 43.77 & 48.18
& 43.36 & 47.34 \\

PyramidKV
& 44.36 & 47.65
& 43.06 & 45.96
& 42.71 & 47.70
& 42.11 & 46.98 \\

CAKE
& 46.30 & 47.97
& 44.73 & 46.66
& 44.70 & 48.52
& 44.49 & 47.51 \\

HeadKV
& 44.11 & 47.05
& 44.10 & 46.41
& 44.21 & 48.42
& 44.02 & 47.50 \\

AdaKV
& 44.45 & 47.94
& 43.75 & 46.38
& 43.68 & 48.19
& 43.30 & 47.33 \\

\textbf{CompressKV}
& \textbf{46.71} & \textbf{48.24}
& \textbf{45.43} & \textbf{46.96}
& \textbf{45.37} & \textbf{48.69}
& \textbf{44.73} & \textbf{47.78} \\

\bottomrule
\end{tabular}
\end{table}

\subsection{Evaluation on Needle In A Haystack}
\begin{figure}[t]
    \centering
    \includegraphics[width=0.92\columnwidth]{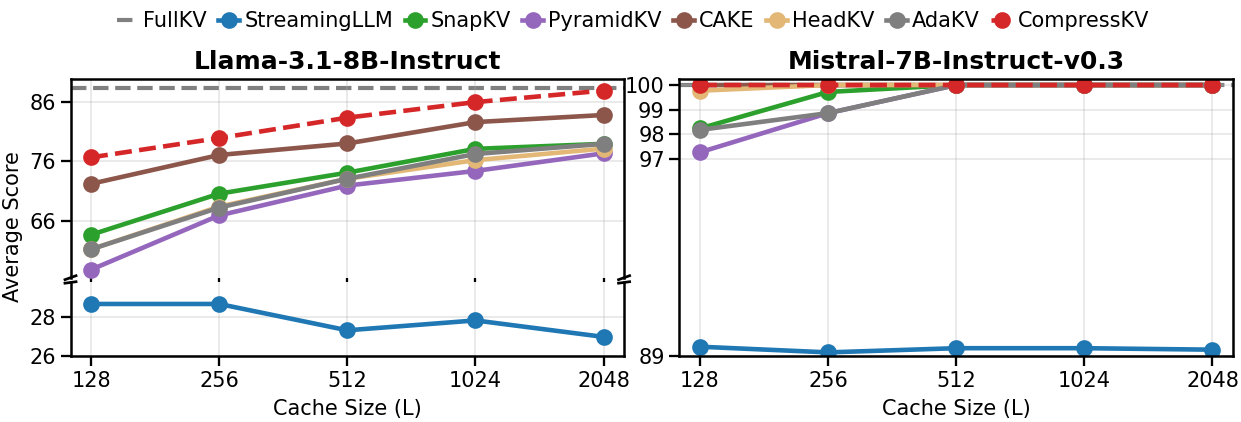}
    \caption{Average performance on the NIAH benchmark under different KV cache budget settings, in comparison with baseline methods.}
    \label{fig:needle_horizontal_bar}
\end{figure}

Figure~\ref{fig:needle_horizontal_bar} presents average Needle-in-a-Haystack performance across KV budgets for Llama-3.1-8B-Instruct (8K–128K context) and Mistral-7B-Instruct-v0.3 (2K–32K)—showing CompressKV consistently surpasses competing methods at every budget. On Mistral-7B-Instruct-v0.3, CompressKV, HeadKV, and CAKE achieve near lossless compression with as few as 256 KV budget, highlighting their robustness. On Llama-3.1-8B-Instruct, AdaKV and HeadKV also underperform at low budgets, while CompressKV achieves nearly lossless performance at a 2048 KV budget (5\% of the full cache) and still retains 90\% of the original performance with only 256 KV budget (0.7\% capacity). Together with the LongBench evaluation, these results show that CompressKV preserves general LLM performance across diverse long-context tasks while delivering efficient KV-cache compression. 
\subsection{Semantic Retrieval Heads: Causal Ablation and Head-Agnostic Gains}
Following the masking-based causal test of \cite{wu2024retrievalheadmechanisticallyexplains}, we conduct targeted ablations on Mistral-7B-Instruct-v0.3.  Specifically, we mask the top-$k$ heads ($k \in \{10,20,30\}$) on the NIAH benchmark. Table~\ref{tab:trh_srh_comparison}~(a) reports the resulting performance drop and compares Semantic Retrieval Heads against traditional Retrieval Heads (TRH). While masking TRH leads to only a minor degradation, masking even a small subset of Semantic Retrieval Heads yields a substantial drop in retrieval accuracy and markedly increases hallucinations, highlighting their critical role in faithful retrieval and localization of supporting evidence. CompressKV is compatible with heterogeneous head definitions. Table~\ref{tab:trh_srh_comparison}~(b) compares CompressKV using TRH vs.\ SRH on Mistral-7B-Instruct-v0.3 under a fixed per-layer KV budget of 256 tokens. SRH yields a modest yet consistent average gain over TRH (+0.24). Moreover, even with TRH and without dynamic budget allocation, CompressKV still surpasses most representative baselines (Table~\ref{tab:longbench_avg}), evidencing more precise salient-token selection.
\begin{table}[t]
\centering
\caption{
Comparison between traditional Retrieval Heads (TRH) and Semantic Retrieval Heads (SRH).
(a) Performance drop after masking top-$k$ heads on NIAH.
(b) LongBench average score when CompressKV uses TRH or SRH under the same KV-cache budget.
Darker cells in (a) indicate larger drops.
}
\label{tab:trh_srh_comparison}

\begin{minipage}[t]{0.56\columnwidth}
\centering
\begin{tabular}{lccc}
\toprule
Heads & Top-10 & Top-20 & Top-30 \\
\midrule
\textbf{TRH}
& \cellcolor{orange!2} 1.02
& \cellcolor{orange!8} 7.67
& \cellcolor{orange!13} 13.30 \\
\textbf{SRH}
& \cellcolor{orange!25} 24.55
& \cellcolor{orange!73} {\color{white}72.56}
& \cellcolor{orange!74} {\color{white}73.81} \\
\bottomrule
\end{tabular}

\smallskip
{\footnotesize (a) Causal masking on NIAH.}
\end{minipage}
\hfill
\begin{minipage}[t]{0.40\columnwidth}
\centering
\begin{tabular}{@{}lcc@{}}
\toprule
Heads & Avg. & $\Delta$ \\
\midrule
\textbf{TRH} & 44.72 & 0.00 \\
\textbf{SRH} & \textbf{44.96} & +0.24 \\
\bottomrule
\end{tabular}

\smallskip
{\footnotesize (b) LongBench average score.}
\end{minipage}
\end{table}

\subsection{Memory and Latency under Long-Context Scaling}
\label{hardware_performance}
\begin{figure}[htbp]
    \centering
    \includegraphics[width=0.92\columnwidth]{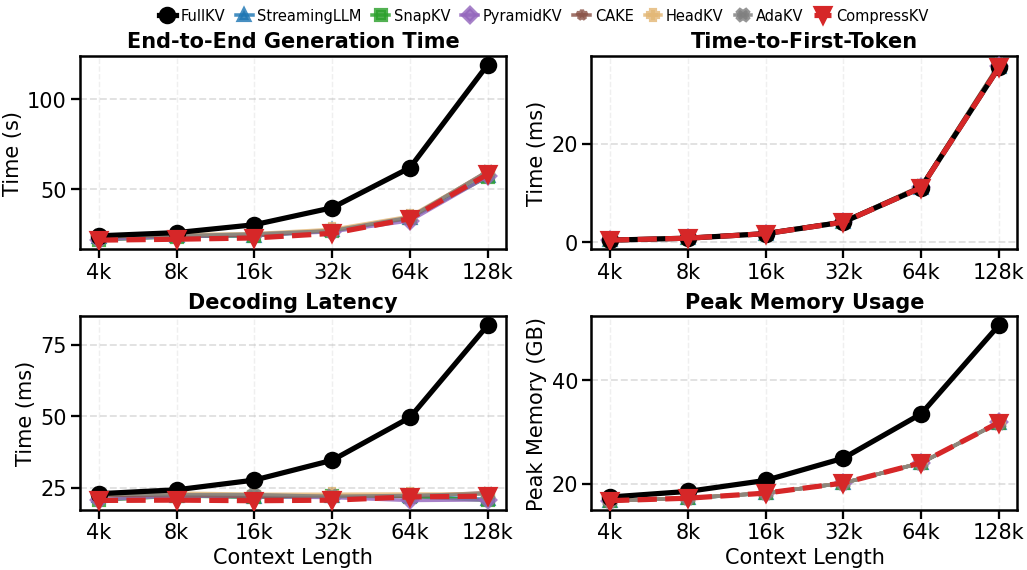}
    \caption{Comprehensive evaluation of inference efficiency on a single NVIDIA A100 GPU.}
    \label{fig:memory_latency_ctx}

\end{figure}

We evaluate end-to-end latency, decoding latency, time  to first token, peak GPU memory, and throughput on Llama-3.1-8B-Instruct with FlashAttention-2~\cite{dao2023flashattention2fasterattentionbetter} on a single NVIDIA A100, sweeping context length from 4K to 128K with a fixed generation length of 1024. We compare CompressKV with a full-cache baseline and six KV-cache eviction methods under a 1024-token KV budget (except full cache). Figure~\ref{fig:memory_latency_ctx} shows that end-to-end latency and time-to-first-token increase with context length for all methods, while eviction-based approaches keep decoding latency nearly constant; in contrast, full-cache decoding latency grows with context length. Under the fixed KV budget, eviction methods have similar peak memory, whereas full cache uses substantially more memory at long contexts. 
\subsection{Ablation Studies}
To evaluate the effectiveness of each part in CompressKV, we conduct a series of ablation studies on the LongBench benchmark using Mistral-7B-Instruct-v0.3 with a fixed KV cache budget of 256.
\paragraph{Token selection and layer-wise cache allocation.}
We ablate SRHs--driven token selection and layer-aware budget allocation on Table~\ref{tab:compresskv_ablation_combined}~(a). Adding our selection to SnapKV improves accuracy; adding layer-aware allocation yields further gains—both components are complementary. 
\paragraph{Number of selected heads per layer.}
We sweep SRH per layer from 2 to 24 (Table~\ref{tab:compresskv_ablation_combined}~(b)). Accuracy peaks at 4 and saturates thereafter (Top-6: $-0.17$; Top-12: $0.00$), with Top-24 slightly worse; thus, 4 heads per layer suffice.
\begin{table}[t]
\centering
\caption{
Component analysis on Mistral-7B-Instruct-v0.3 under a fixed KV-cache budget of 256.
}
\label{tab:compresskv_ablation_combined}
{

\begin{minipage}[c]{0.34\linewidth}
\centering
{

\begin{tabular}{lc}
\toprule
Method & Acc. (\%) \\
\midrule
SnapKV & 43.76 \\
\rowcolor{blue!6}
+ SRH Selection & 44.96 \\
\rowcolor{blue!10}
+ SRH + Layer Alloc. & \textbf{45.43} \\
\bottomrule
\end{tabular}
}
\smallskip

{\footnotesize (a) Contribution of each component.}
\end{minipage}
\hfill
\begin{minipage}[c]{0.60\linewidth}
\centering
{

\begin{tabular}{lcc}
\toprule
SRHs per Layer & Mean Acc. (\%) & $\Delta$ vs. Top-4 \\
\midrule
Top-2  & 44.33 & $-0.63$ \\
\rowcolor{blue!8}
Top-4  & \textbf{44.96} & 0.00 \\
Top-6  & 44.79 & $-0.17$ \\
Top-12 & 44.96 & 0.00 \\
Top-24 & 44.30 & $-0.66$ \\
\bottomrule
\end{tabular}
}
\smallskip

{\footnotesize (b) Effect of the number of SRHs per layer.}
\end{minipage}
}
\end{table}

\subsection{Orthogonal to Prior Efficiency Methods}
\label{sec:orthogonal_integration}
\begin{figure}[!ht]
    \centering
    \includegraphics[width=0.92\columnwidth]{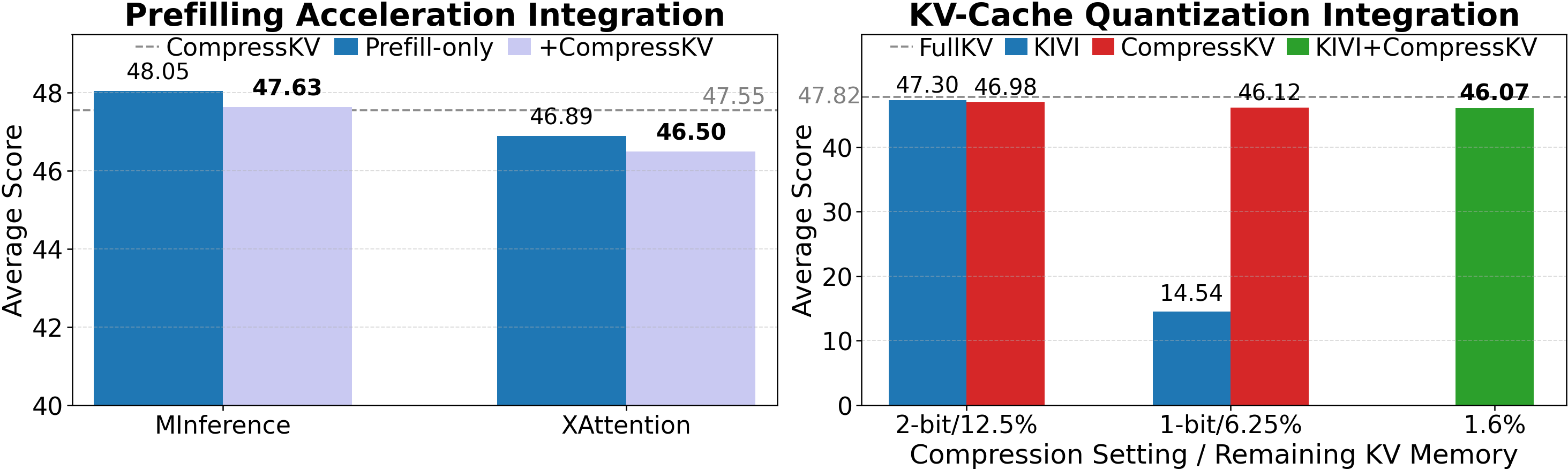}
    \caption{
    Integration of CompressKV with existing efficiency techniques on Mistral-7B-Instruct-v0.3.
    Left: integration with prefilling-stage accelerators; the dashed line denotes standalone CompressKV.
    Right: integration with KV-cache quantization; the dashed line denotes 16-bit FullKV.
    }
    \label{fig:integration_combined}
\end{figure}

\paragraph{With Prefilling Acceleration.}
CompressKV can be integrated with prefilling-stage accelerators such as MInference~\cite{jiang2024minference} and XAttention~\cite{xu2025xattention}, as they target prefilling cost while CompressKV targets decoding-stage KV-cache memory. 
We conduct the integration experiments on Mistral-7B-Instruct-v0.3 under a 2048-token per-layer KV-cache budget. 
As shown in Figure~\ref{fig:integration_combined}, the combined variants maintain accuracy close to the prefilling-only baselines while further reducing decoding memory.
\paragraph{With KV-Cache Quantization.}
CompressKV also complements KV-cache quantization methods such as KIVI~\cite{liu2024kivi}: KIVI reduces KV precision, whereas CompressKV prunes less critical tokens while preserving full-precision KV entries. 
On Mistral-7B-Instruct-v0.3 with a 2048-token per-layer KV-cache budget, Figure~\ref{fig:integration_combined} shows that 2-bit KIVI slightly outperforms CompressKV at comparable memory usage, but degrades sharply under 1-bit quantization. 
In contrast, CompressKV remains robust, and combining it with 2-bit KIVI further reduces KV memory to about $1.6\%$ of the 16-bit full-cache baseline while maintaining strong accuracy.

\begin{figure}[!htbp]
    \centering
    \includegraphics[width=0.92\columnwidth]{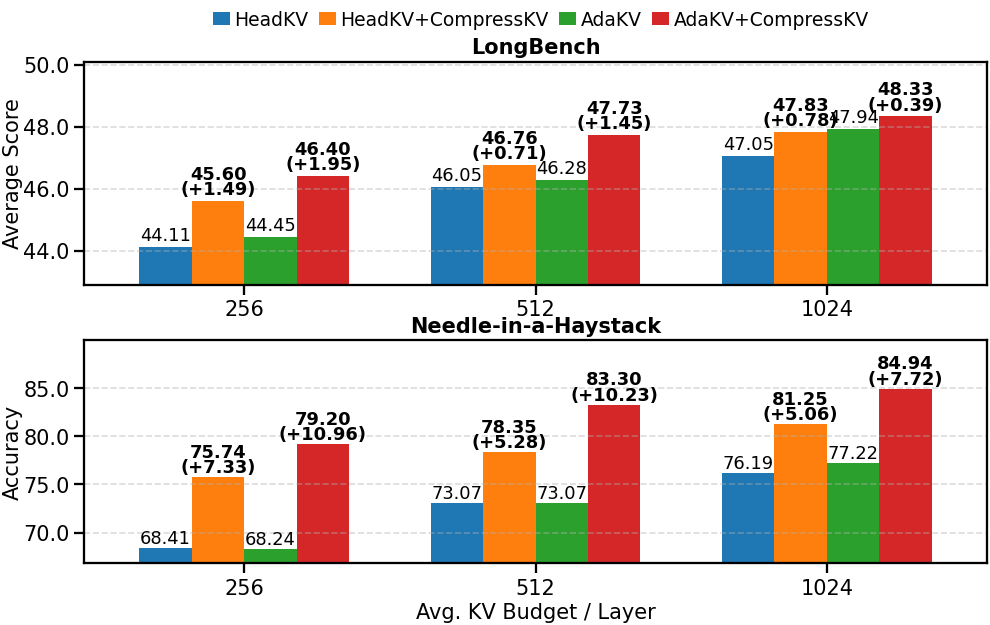}
    \caption{Integration of CompressKV with head-level allocation methods on Llama-3.1-8B-Instruct}
    \label{fig:headlevel_score_delta}
\end{figure}

\paragraph{With Head-Level Allocation.}
CompressKV can also be combined with head-level budget allocation methods such as HeadKV~\cite{fu2024headsmatterheadlevelkv} and AdaKV~\cite{feng2025adakvoptimizingkvcache}. 
We evaluate these integrations on LLaMA-3.1-8B-Instruct under different KV-cache budgets. 
Integrating our token selection with HeadKV yields HeadCompressKV, while combining our token selection with error-aware layer-wise allocation on AdaKV yields AdaCompressKV. 
As shown in Figure~\ref{fig:headlevel_score_delta}, both variants consistently improve performance across KV-cache budgets, achieving gains of up to nearly 2 points on LongBench and 11 points on Needle-in-a-Haystack under tight memory.
\subsection{Head Visualization}
\label{sec:head_visualization}
\begin{figure}[!htbp]
    \centering
    \includegraphics[width=0.92\columnwidth]{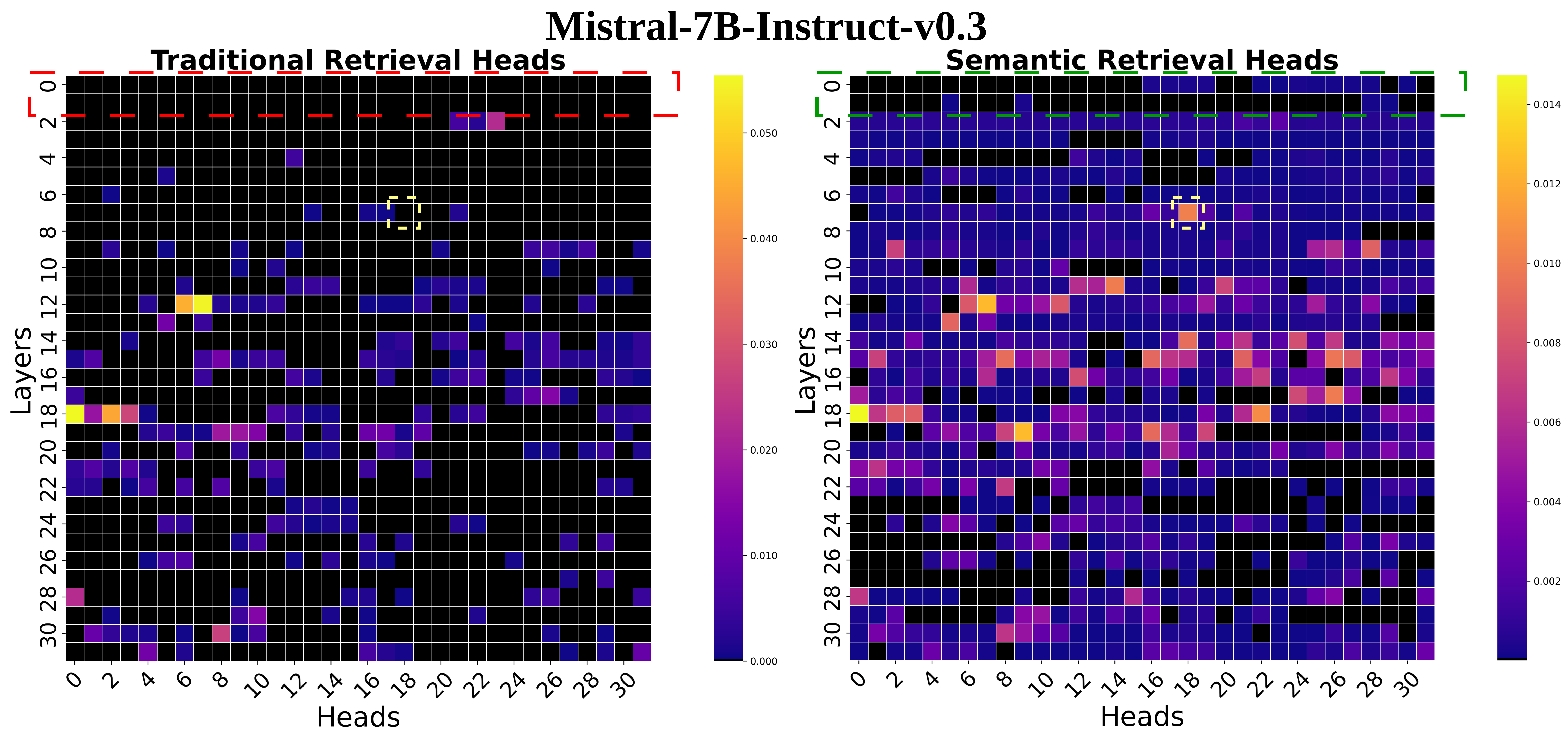}
    \caption{Head visualization for Mistral-7B-Instruct-v0.3. Left: Traditional Retrieval Heads. Right: Semantic Retrieval Heads identified.}
    \label{fig:mistral-head-vis}
\end{figure}

In Figures~\ref{fig:mistral-head-vis}, we present a comparison between traditional Retrieval Heads and Semantic Retrieval Heads identified using Mistral-7B-Instruct-v0.3. All scores are L1-normalized across the attention head importance distributions. Unlike traditional methods that require exact top-$k$ attention hits, our approach aggregates scores over entire answer spans, capturing heads that contribute semantically relevant context even when they never achieve top-1 attention for individual tokens. For instance, as shown in Figure~\ref{fig:mistral-head-vis}, layers 0 and 1 of the Mistral model have zero scores for all heads using the traditional method, whereas our approach successfully identifies heads of lower yet meaningful importance.
\section{Conclusion}
We presented CompressKV, a KV-cache compression framework for GQA-based LLMs that improves the resource--performance trade-off of long-context inference. CompressKV identifies Semantic Retrieval Heads to avoid streaming-head-dominated token eviction and uses offline layer-wise eviction errors to allocate cache budgets adaptively across layers. Experiments on LongBench and Needle-in-a-Haystack across multiple models and cache budgets show that CompressKV consistently preserves accuracy under tight KV-cache memory constraints. These results demonstrate that retrieval-aware token selection and error-aware allocation provide an effective path toward more memory-efficient and sustainable long-context LLM inference.

\section*{Acknowledgement}
This work is funded by the European Union - European Research Council (ERC) Starting Grant - Project-ID 101219243. Views and opinions expressed are however those of the author(s) only and do not necessarily reflect those of the European Union or the European Research Council. Neither the European Union nor the granting authority can be held responsible for them.

\section*{Declaration on Generative AI}
During the preparation of this work, the authors used ChatGPT (OpenAI) to assist with grammar improvement, language polishing, and manuscript editing. After using this tool, the authors carefully reviewed and revised all content and take full responsibility for the accuracy, originality, and integrity of the publication.
\bibliography{sample-ceur}

@inproceedings{
jiang2024minference,
title={{MI}nference 1.0: Accelerating Pre-filling for Long-Context {LLM}s via Dynamic Sparse Attention},
author={Huiqiang Jiang and YUCHENG LI and Chengruidong Zhang and Qianhui Wu and Xufang Luo and Surin Ahn and Zhenhua Han and Amir H. Abdi and Dongsheng Li and Chin-Yew Lin and Yuqing Yang and Lili Qiu},
booktitle={The Thirty-eighth Annual Conference on Neural Information Processing Systems},
year={2024},
url={https://openreview.net/forum?id=fPBACAbqSN}
}

@inproceedings{
xu2025xattention,
title={{XA}ttention: Block Sparse Attention with Antidiagonal Scoring},
author={Ruyi Xu and Guangxuan Xiao and Haofeng Huang and Junxian Guo and Song Han},
booktitle={Forty-second International Conference on Machine Learning},
year={2025},
url={https://openreview.net/forum?id=KG6aBfGi6e}
}

@inproceedings{
jingcun2025,
title={Basis Sharing: Cross-Layer Parameter Sharing for Large Language Model Compression},
author={Jingcun Wang and Yu-Guang Chen and Ing-Chao Lin and Bing Li and Grace Li Zhang},
booktitle={The Thirteenth International Conference on Learning Representations},
year={2025},
url={https://openreview.net/forum?id=gp32jvUquq}
}

@misc{feng2025adakvoptimizingkvcache,
      title={Ada-KV: Optimizing KV Cache Eviction by Adaptive Budget Allocation for Efficient LLM Inference}, 
      author={Yuan Feng and Junlin Lv and Yukun Cao and Xike Xie and S. Kevin Zhou},
      year={2025},
      eprint={2407.11550},
      archivePrefix={arXiv},
      primaryClass={cs.CL},
      url={https://arxiv.org/abs/2407.11550}, 
}

@inproceedings{fu2024headsmatterheadlevelkv,
      title={Not All Heads Matter: A Head-Level KV Cache Compression Method with Integrated Retrieval and Reasoning}, 
      author={Yu Fu and Zefan Cai and Abedelkadir Asi and Wayne Xiong and Yue Dong and Wen Xiao},
      booktitle={The Twelfth International Conference on Learning Representations},
        year={2025},
      url={https://arxiv.org/abs/2410.19258}, 
}

@inproceedings{tang2024razorattentionefficientkvcache,
      title={RazorAttention: Efficient KV Cache Compression Through Retrieval Heads}, 
      author={Hanlin Tang and Yang Lin and Jing Lin and Qingsen Han and Shikuan Hong and Yiwu Yao and Gongyi Wang},
    booktitle={The Twelfth International Conference on Learning Representations},
    year={2025},
    url={https://arxiv.org/abs/2407.15891}
}

@inproceedings{
xiao2024efficientstreaminglanguagemodels,
title={Efficient Streaming Language Models with Attention Sinks},
author={Guangxuan Xiao and Yuandong Tian and Beidi Chen and Song Han and Mike Lewis},
booktitle={The Twelfth International Conference on Learning Representations},
year={2024},
url={https://openreview.net/forum?id=NG7sS51zVF}
}

@inproceedings{li2024snapkvllmknowslooking,
    title={Snap{KV}: {LLM} Knows What You are Looking for Before Generation},
    author={Yuhong Li and Yingbing Huang and Bowen Yang and Bharat Venkitesh and Acyr Locatelli and Hanchen Ye and Tianle Cai and Patrick Lewis and Deming Chen},
    booktitle={The Thirty-eighth Annual Conference on Neural Information Processing Systems},
    year={2024},
    url={https://openreview.net/forum?id=poE54GOq2l}
}

@misc{cai2025pyramidkvdynamickvcache,
      title={PyramidKV: Dynamic KV Cache Compression based on Pyramidal Information Funneling}, 
      author={Zefan Cai and Yichi Zhang and Bofei Gao and Yuliang Liu and Yucheng Li and Tianyu Liu and Keming Lu and Wayne Xiong and Yue Dong and Junjie Hu and Wen Xiao},
      year={2025},
      eprint={2406.02069},
      archivePrefix={arXiv},
      primaryClass={cs.CL},
      url={https://arxiv.org/abs/2406.02069}, 
}

@inproceedings{
qin2025cakecascadingadaptivekv,
title={{CAKE}: Cascading and Adaptive {KV} Cache Eviction with Layer Preferences},
author={Ziran Qin and Yuchen Cao and Mingbao Lin and Wen Hu and Shixuan Fan and Ke Cheng and Weiyao Lin and Jianguo Li},
booktitle={The Thirteenth International Conference on Learning Representations},
year={2025},
url={https://openreview.net/forum?id=EQgEMAD4kv}
}

@inproceedings{bai2024longbenchbilingualmultitaskbenchmark,
    title = "{L}ong{B}ench: A Bilingual, Multitask Benchmark for Long Context Understanding",
    author = "Bai, Yushi  and
      Lv, Xin  and
      Zhang, Jiajie  and
      Lyu, Hongchang  and
      Tang, Jiankai  and
      Huang, Zhidian  and
      Du, Zhengxiao  and
      Liu, Xiao  and
      Zeng, Aohan  and
      Hou, Lei  and
      Dong, Yuxiao  and
      Tang, Jie  and
      Li, Juanzi",
    booktitle = "Proceedings of the 62nd Annual Meeting of the Association for Computational Linguistics (Volume 1: Long Papers)",
    year = "2024",
    url = "https://aclanthology.org/2024.acl-long.172/",
}

@inproceedings{
    wu2024retrievalheadmechanisticallyexplains,
    title={Retrieval Head Mechanistically Explains Long-Context Factuality},
    author={Wenhao Wu and Yizhong Wang and Guangxuan Xiao and Hao Peng and Yao Fu},
    booktitle={The Thirteenth International Conference on Learning Representations},
    year={2025},
    url={https://openreview.net/forum?id=EytBpUGB1Z}
}

@misc{gkamradt2024llmtest,
  author       = {Kamradt, Greg},
  title        = {NeedleInAHaystack},
  howpublished = {\url{https://github.com/gkamradt/LLMTest_NeedleInAHaystack}},
  note         = {Accessed: 2025-07-13},
  year         = {2023}
}

@misc{olsson2022incontextlearninginductionheads,
      title={In-context Learning and Induction Heads}, 
      author={Catherine Olsson and Nelson Elhage and Neel Nanda and Nicholas Joseph and Nova DasSarma and Tom Henighan and Ben Mann and Amanda Askell and Yuntao Bai and Anna Chen and Tom Conerly and Dawn Drain and Deep Ganguli and Zac Hatfield-Dodds and Danny Hernandez and Scott Johnston and Andy Jones and Jackson Kernion and Liane Lovitt and Kamal Ndousse and Dario Amodei and Tom Brown and Jack Clark and Jared Kaplan and Sam McCandlish and Chris Olah},
      year={2022},
      eprint={2209.11895},
      archivePrefix={arXiv},
      primaryClass={cs.LG},
      url={https://arxiv.org/abs/2209.11895}, 
}

@inproceedings{ren2024identifyingsemanticinductionheads,
    title = "Identifying Semantic Induction Heads to Understand In-Context Learning",
    author = "Ren, Jie  and
      Guo, Qipeng  and
      Yan, Hang  and
      Liu, Dongrui  and
      Zhang, Quanshi  and
      Qiu, Xipeng  and
      Lin, Dahua",
    booktitle = "Findings of the Association for Computational Linguistics: ACL 2024",
    year = "2024",

    url = "https://aclanthology.org/2024.findings-acl.412/",
}

@inproceedings{
    todd2024functionvectorslargelanguage,
    title={Function Vectors in Large Language Models},
    author={Eric Todd and Millicent Li and Arnab Sen Sharma and Aaron Mueller and Byron C Wallace and David Bau},
    booktitle={The Twelfth International Conference on Learning Representations},
    year={2024},
    url={https://openreview.net/forum?id=AwyxtyMwaG}
}

@misc{yin2025attentionheadsmatterincontext,
      title={Which Attention Heads Matter for In-Context Learning?}, 
      author={Kayo Yin and Jacob Steinhardt},
      year={2025},
      eprint={2502.14010},
      archivePrefix={arXiv},
      primaryClass={cs.LG},
      url={https://arxiv.org/abs/2502.14010}, 
}

@misc{jiang2024clipdinovisualencoders,
      title={From CLIP to DINO: Visual Encoders Shout in Multi-modal Large Language Models}, 
      author={Dongsheng Jiang and Yuchen Liu and Songlin Liu and Jin'e Zhao and Hao Zhang and Zhen Gao and Xiaopeng Zhang and Jin Li and Hongkai Xiong},
      year={2024},
      eprint={2310.08825},
      archivePrefix={arXiv},
      primaryClass={cs.CV},
      url={https://arxiv.org/abs/2310.08825}, 
}

@misc{grattafiori2024llama3herdmodels,
      title={The Llama 3 Herd of Models}, 
      author={Dubey, Abhimanyu and Jauhri, Abhinav and Pandey, Abhinav and Kadian, Abhishek and Al-Dahle, Ahmad and Letman, Aiesha and Mathur, Akhil and Schelten, Alan and Yang, Amy and Fan, Angela and others},
      year={2024},
      eprint={2407.21783},
      archivePrefix={arXiv},
      primaryClass={cs.AI},
      url={https://arxiv.org/abs/2407.21783}, 
}

@inproceedings{
dao2023flashattention2fasterattentionbetter,
title={FlashAttention-2: Faster Attention with Better Parallelism and Work Partitioning},
author={Tri Dao},
booktitle={The Twelfth International Conference on Learning Representations},
year={2024},
url={https://openreview.net/forum?id=mZn2Xyh9Ec}
}

@misc{openai2024gpt4technicalreport,
      title={GPT-4 Technical Report}, 
      author={Achiam, Josh and Adler, Steven and Agarwal, Sandhini and Ahmad, Lama and Akkaya, Ilge and Aleman, Florencia Leoni and Almeida, Diogo and Altenschmidt, Janko and Altman, Sam and Anadkat, Shyamal and others},
      year={2024},
      eprint={2303.08774},
      archivePrefix={arXiv},
      primaryClass={cs.CL},
      url={https://arxiv.org/abs/2303.08774}, 
}

@misc{qwen2025qwen25technicalreport,
      title={Qwen2.5 Technical Report}, 
      author={Hui, Binyuan and Yang, Jian and Cui, Zeyu and Yang, Jiaxi and Liu, Dayiheng and Zhang, Lei and Liu, Tianyu and Zhang, Jiajun and Yu, Bowen and Lu, Keming and others},
      year={2025},
      eprint={2412.15115},
      archivePrefix={arXiv},
      primaryClass={cs.CL},
      url={https://arxiv.org/abs/2412.15115}, 
}

@techreport{anthropic_claude3_2024,
    author = {Anthropic},
    title = {The Claude 3 Model Family: Opus, Sonnet, Haiku},
    institution ={Anthropic},
    year         = {2024},
    url          = {https://www-cdn.anthropic.com/de8ba9b01c9ab7cbabf5c33b80b7bbc618857627/Model_Card_Claude_3.pdf},
    note         = {Accessed: 2024-07-09}
}

@inproceedings{yang2024pyramidinferpyramidkvcache,
  title={PyramidInfer: Pyramid KV Cache Compression for High-throughput LLM Inference},
  author={Yang, Dongjie and Han, Xiaodong and Gao, Yan and Hu, Yao and Zhang, Shilin and Zhao, Hai},
  booktitle={Findings of the Association for Computational Linguistics ACL 2024},
  pages={3258--3270},
  year={2024}
}

@inproceedings{
    ge2024modeltellsdiscardadaptive,
    title={Model Tells You What to Discard: Adaptive KV Cache Compression for LLMs}, 
    author={Suyu Ge and Yunan Zhang and Liyuan Liu and Minjia Zhang and Jiawei Han and Jianfeng Gao},
    booktitle={The Thirteenth International Conference on Learning Representations},
    year={2024},
    url={https://openreview.net/pdf?id=uNrFpDPMyo}
}

@inproceedings{liu2023scissorhandsexploitingpersistenceimportance,
    title={Scissorhands: Exploiting the Persistence of Importance Hypothesis for {LLM} {KV} Cache Compression at Test Time},
    author={Zichang Liu and Aditya Desai and Fangshuo Liao and Weitao Wang and Victor Xie and Zhaozhuo Xu and Anastasios Kyrillidis and Anshumali Shrivastava},
    booktitle={Thirty-seventh Conference on Neural Information Processing Systems},
    year={2023},
    url={https://openreview.net/forum?id=JZfg6wGi6g}
}

@inproceedings{
    zhang2023h2oheavyhitteroracleefficient,
    title={H2O: Heavy-Hitter Oracle for Efficient Generative Inference of Large Language Models},
    author={Zhenyu Zhang and Ying Sheng and Tianyi Zhou and Tianlong Chen and Lianmin Zheng and Ruisi Cai and Zhao Song and Yuandong Tian and Christopher Re and Clark Barrett and Zhangyang Wang and Beidi Chen},
    booktitle={Thirty-seventh Conference on Neural Information Processing Systems},
    year={2023},
    url={https://openreview.net/forum?id=RkRrPp7GKO}
}

@inproceedings{han2024lminfinitezeroshotextremelength,
  title={LM-Infinite: Zero-Shot Extreme Length Generalization for Large Language Models},
  author={Han, Chi and Wang, Qifan and Peng, Hao and Xiong, Wenhan and Chen, Yu and Ji, Heng and Wang, Sinong},
  booktitle={Proceedings of the 2024 Conference of the North American Chapter of the Association for Computational Linguistics: Human Language Technologies (Volume 1: Long Papers)},
  pages={3991--4008},
  year={2024}
}

@inproceedings{oren2024transformersmultistaternns,
  title={Transformers are Multi-State RNNs},
  author={Oren, Matanel and Hassid, Michael and Yarden, Nir and Adi, Yossi and Schwartz, Roy},
  booktitle={Proceedings of the 2024 Conference on Empirical Methods in Natural Language Processing},
  pages={18724--18741},
  year={2024}
}

@inproceedings{
    wan2025d2odynamicdiscriminativeoperations,
    title={D2O: Dynamic Discriminative Operations for Efficient Long-Context Inference of Large Language Models},
    author={Zhongwei Wan and Xinjian Wu and Yu Zhang and Yi Xin and Chaofan Tao and Zhihong Zhu and Xin Wang and Siqi Luo and Jing Xiong and Longyue Wang and Mi Zhang},
    booktitle={The Thirteenth International Conference on Learning Representations},
    year={2025},
    url={https://openreview.net/forum?id=HzBfoUdjHt}
}

@inproceedings{ainslie2023gqatraininggeneralizedmultiquery,
  title={GQA: Training Generalized Multi-Query Transformer Models from Multi-Head Checkpoints},
  author={Ainslie, Joshua and Lee-Thorp, James and de Jong, Michiel and Zemlyanskiy, Yury and Lebron, Federico and Sanghai, Sumit},
  booktitle={The 2023 Conference on Empirical Methods in Natural Language Processing},
  year={2023},
 url={https://openreview.net/forum?id=hmOwOZWzYE}
}

@misc{zheng2024attentionheadslargelanguage,
      title={Attention Heads of Large Language Models: A Survey}, 
      author={Zifan Zheng and Yezhaohui Wang and Yuxin Huang and Shichao Song and Mingchuan Yang and Bo Tang and Feiyu Xiong and Zhiyu Li},
      year={2024},
      eprint={2409.03752},
      archivePrefix={arXiv},
      primaryClass={cs.CL},
      url={https://arxiv.org/abs/2409.03752}, 
}

@inproceedings{
    kwon2022fastposttrainingpruningframework,
    title={A Fast Post-Training Pruning Framework for Transformers},
    author={Woosuk Kwon and Sehoon Kim and Michael W. Mahoney and Joseph Hassoun and Kurt Keutzer and Amir Gholami},
    booktitle={Advances in Neural Information Processing Systems},
    editor={Alice H. Oh and Alekh Agarwal and Danielle Belgrave and Kyunghyun Cho},
    year={2022},
    url={https://openreview.net/forum?id=0GRBKLBjJE}
}

@inproceedings{
xiao2024duoattentionefficientlongcontextllm,
title={DuoAttention: Efficient Long-Context {LLM} Inference with Retrieval and Streaming Heads},
author={Guangxuan Xiao and Jiaming Tang and Jingwei Zuo and junxian guo and Shang Yang and Haotian Tang and Yao Fu and Song Han},
booktitle={The Thirteenth International Conference on Learning Representations},
year={2025},
url={https://openreview.net/forum?id=cFu7ze7xUm}
}

@inproceedings{
liu2024kivi,
title={{KIVI}: A Tuning-Free Asymmetric 2bit Quantization for {KV} Cache},
author={Zirui Liu and Jiayi Yuan and Hongye Jin and Shaochen Zhong and Zhaozhuo Xu and Vladimir Braverman and Beidi Chen and Xia Hu},
booktitle={Forty-first International Conference on Machine Learning},
year={2024},
url={https://openreview.net/forum?id=L057s2Rq8O}
}

@misc{kang2024gearefficientkvcache,
      title={GEAR: An Efficient KV Cache Compression Recipe for Near-Lossless Generative Inference of LLM}, 
      author={Hao Kang and Qingru Zhang and Souvik Kundu and Geonhwa Jeong and Zaoxing Liu and Tushar Krishna and Tuo Zhao},
      year={2024},
      eprint={2403.05527},
      archivePrefix={arXiv},
      primaryClass={cs.LG},
      url={https://arxiv.org/abs/2403.05527}, 
}

\end{document}